\documentclass{article}
\usepackage{spconf,amsmath,graphicx}
\usepackage{amsmath,amssymb,mathtools,amsfonts}
\usepackage{graphicx,subfig,color,wrapfig,epsfig,epstopdf}
\usepackage{url}
\usepackage{multirow}
\usepackage{listings}
\usepackage{xspace}
\usepackage{cleveref}
\usepackage{INTERSPEECH2019}

\newcommand{\ssgd}{{S-PSGD}\xspace}
\newcommand{\gdpsgd}{{(A)D-PSGD}\xspace}
\newcommand{\dpsgd}{{D-PSGD}\xspace}
\newcommand{\adpsgd}{{AD-PSGD}\xspace}

\newcommand{\rad}{{RAND-PSGD}\xspace}
\newcommand{\done}{{D1D-PSGD}\xspace}

\newcommand{\swbb}{{SWB2000}\xspace}
\newcommand{\imagenet}{{\tt{ImageNet}}\xspace}
\newcommand{\sg}{spectral gap\xspace}
\newcommand{\trans}[1]{{#1}^{\ensuremath{\mathsf{T}}}}

\title{Improving Efficiency in Large-Scale Decentralized Distributed Training}
\name{Wei Zhang, Xiaodong Cui, Abdullah Kayi, Mingrui Liu*, Ulrich Finkler, Brian Kingsbury, \\ George Saon, Youssef Mroueh, Alper Buyuktosunoglu, Payel Das, David Kung, Michael Picheny}
\address{IBM Research, \  \  *University of Iowa}
\email{\{weiz,cuix,kayi,ufinkler,bedk,gsaon,mroueh,alperb,daspa,kung,picheny\}@us.ibm.com, *mingrui-liu@uiowa.edu}
\begin{document}
\ninept
\maketitle
\begin{abstract}
  %% Designing an efficient large-scale distributed learning strategy that has a good convergence behavior and low communication cost is not only theoretically appealing but also practically challenging.
  Decentralized Parallel SGD (D-PSGD) and its asynchronous variant Asynchronous Parallel SGD (\adpsgd) is a family of distributed learning algorithms that have been demonstrated to perform well for large-scale deep learning tasks. One drawback of \gdpsgd is that the \sg of the mixing matrix decreases when the number of learners in the system increases, which hampers convergence. In this paper, we investigate techniques to accelerate \gdpsgd based training by improving the \sg while minimizing the communication cost. %% Specifically, we devise a randomized local averaging scheme to improve the spectral gap of the doubly stochastic mixing matrix in D-PSGD, which leads to faster convergence than that of fixed local averaging. We also investigate a delay-by-one scheme which allows large batch sizes for speedup.
  We demonstrate the effectiveness of our proposed techniques by running experiments on the 2000-hour Switchboard speech recognition task and the ImageNet computer vision task. %% We show that the investigated strategies can significantly improve the training efficiency in term of convergence and communication cost.
  On an IBM P9 supercomputer, our system is able to train an LSTM acoustic model in 2.28 hours with 7.5\% WER on the Hub5-2000 Switchboard (SWB) test set and 13.3\% WER on the CallHome (CH) test set using 64 V100 GPUs and in 1.98 hours with 7.7\% WER on SWB and 13.3\% WER on CH using 128 V100 GPUs, the fastest training time reported to date.
\end{abstract}
\begin{keywords}
distributed training, decentralized SGD, parallel computing, automatic speech recognition, image recognition.
\end{keywords}
\section{Introduction}
\label{sec:intro}

\begin{figure}[t]%[!htpb]
\small
    \centering
    \subfloat[{\swbb Held-out loss}]{{ \includegraphics[width=0.24\textwidth, scale=1.0]{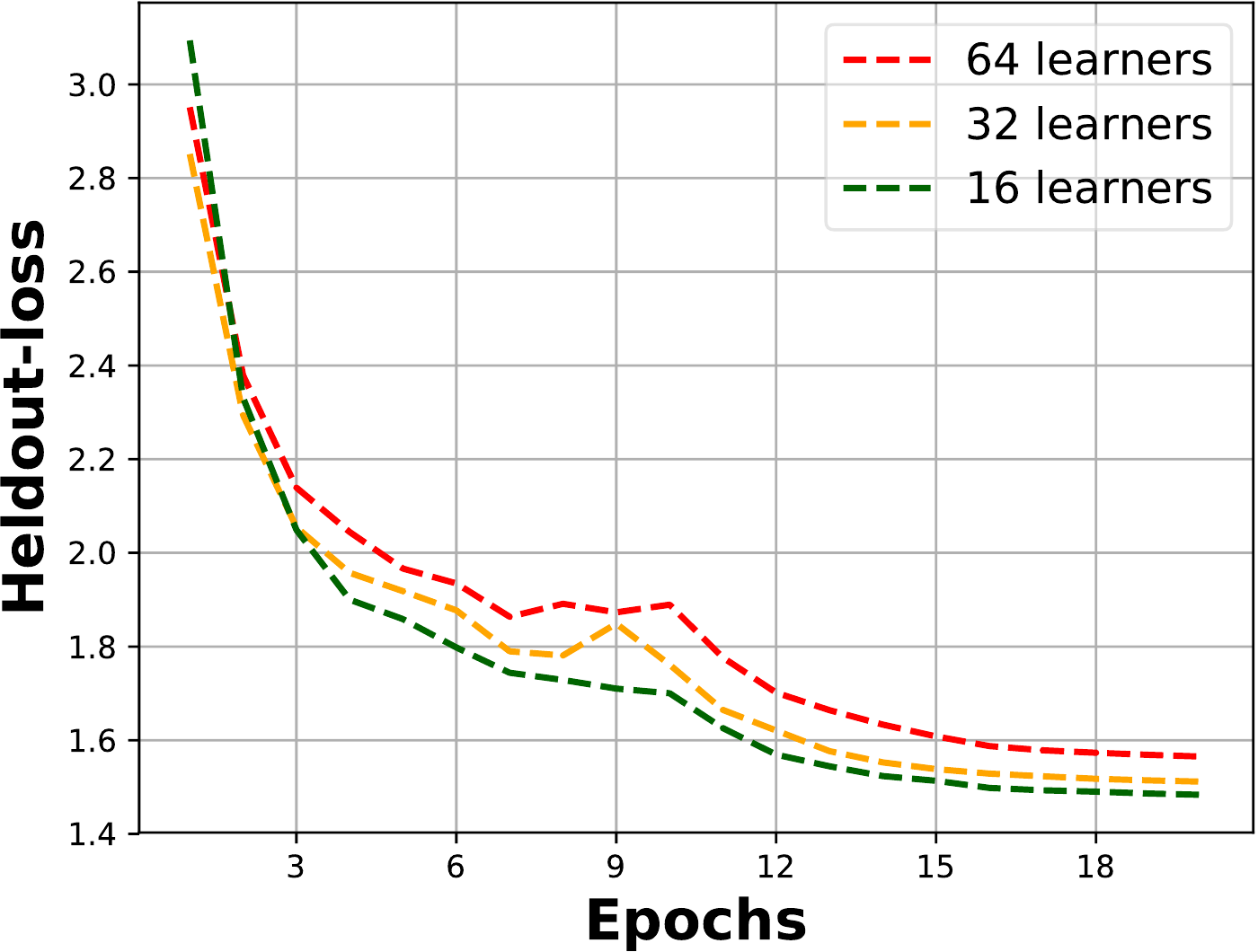} }\label{fig:motivation_swb}}%\hfill
    \subfloat[{ImageNet Top-1 accuracy}]{{ \includegraphics[width=0.24\textwidth, scale=1.0]{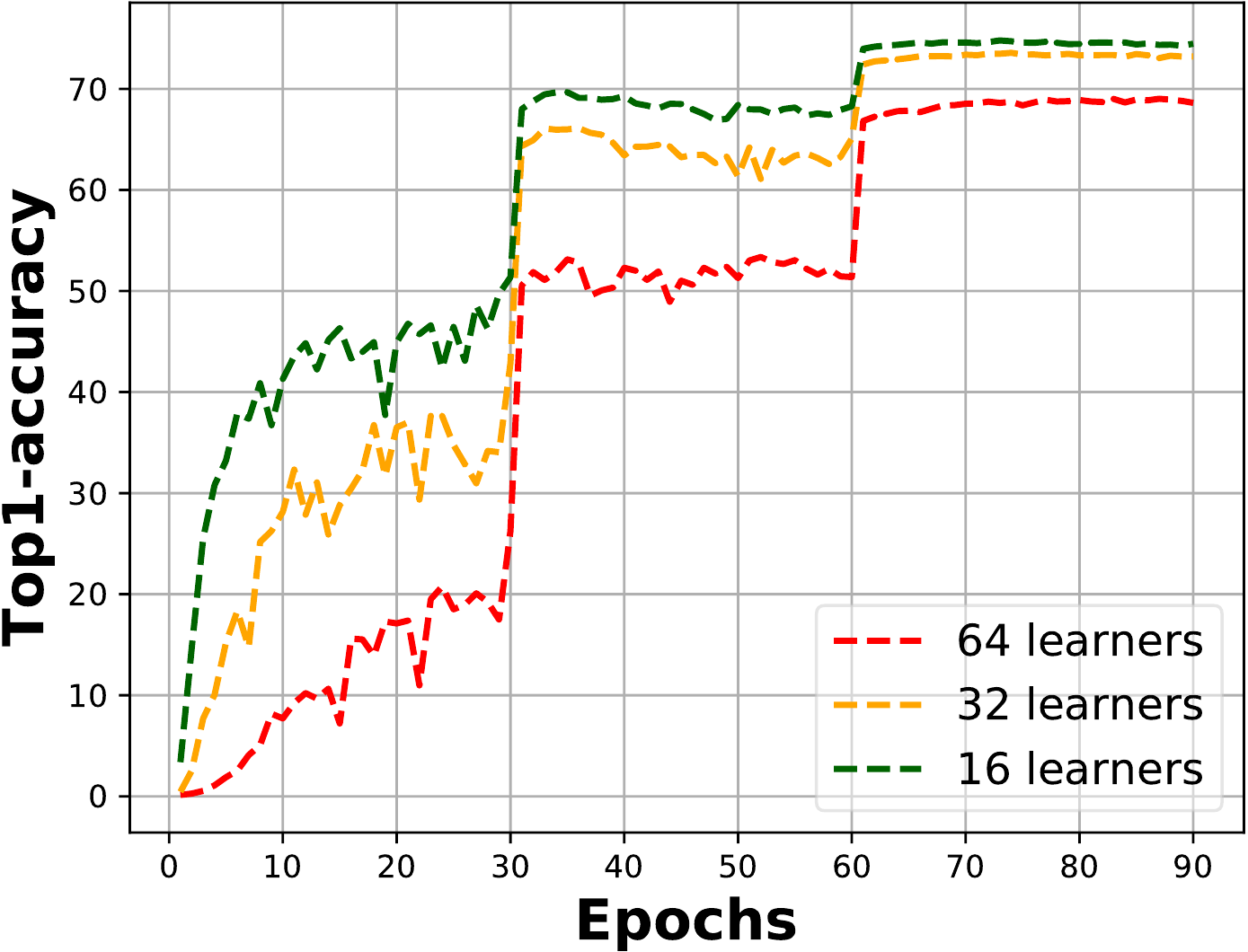} }\label{fig:motivation_imagenet}}
\caption{In AD-PSGD convergence slows down when the number of learners grows. \swbb hyper-parameters are set as prescribed in \cite{interspeech19} with total batch size fixed at 8192. ImageNet hyper-parameters are set as prescribed in \cite{facebook-1hr} with batch size per learner fixed at 32.}
\label{fig:motivation}
\vspace{-0.3cm}
\end{figure}

Large-scale distributed training plays an important role in deep learning to deal with large amounts of training data and models with deep architectures. An efficient distributed training algorithm aims at maximizing the convergence rate while minimizing the communication cost. %% and communication among learners and high task performance of interest (e.g. recognition accuracy for the SWB2000 and ImageNet tasks).
Synchronous Parallel SGD (\ssgd) is the de-facto distributed learning algorithms in practice. Recently, Decentralized Parallel SGD (\dpsgd) \cite{dpsgd} and its asynchronous variant Asynchronous Decentralized Parallel SGD (\adpsgd) \cite{adpsgd} have been applied to a broad variety of deep learning tasks. Compared to \ssgd, \gdpsgd replaces global weight synchronization with model averaging among neighboring learners in a peer-to-peer fashion while achieving the same convergence rate. In \cite{icassp19}, \adpsgd was first applied to automatic speech recognition (ASR) to significantly shorten the acoustic model training time. In \cite{interspeech19}, it was discovered that \gdpsgd can converge with a much larger batch size than \ssgd , which enables a larger degree of parallelism for distributed training. One drawback of \gdpsgd, however, is that when the number of learners grows, it requires more rounds of communication to reach consensus, slowing down convergence. \Cref{fig:motivation} illustrates \gdpsgd convergence curves for the 2000-hour Switchboard (SWB2000) and ImageNet tasks when running with different numbers of learners. It shows when the number of learners increases, the convergence slows down.

In this paper, we investigate techniques to improve large-scale \gdpsgd based training.  In \Cref{sec:background} we formulate the \gdpsgd problem. In \Cref{sec:theory}, we first analyze why the fixed model averaging among neighboring learners, as proposed in the original \gdpsgd, incurs slow convergence to consensus. Based on that, we propose a randomized mixing scheme, Randomization Accelerated Decentralized Parallel SGD (\rad), that can significantly improve the spectral gap of the mixing matrix to improve the convergence to consensus, while maintaining the same communication cost. We further investigate the ``Delay-by-one" Decentralized Parallel SGD (\done) scheme which ensures the weights used to calculate gradients and the weights consensus differ by precisely one iteration of gradients calculation. \done enables the fast speed to reach consensus while maintaining the decentralized training structure so that it can still converge under a larger batch size compared to \ssgd, at the cost of placing a global synchronization in a separate communication thread.

We describe the implementation details in \Cref{sec:design} and present experimental results of \rad and \done and discuss the trade-offs of each design choice in \Cref{sec:results}. We discuss related works in \Cref{sec:related} and conclude with a summary in \Cref{sec:conclusion}.

%This paper makes the following contributions: (1) We identify the connection between number of learners and \sg of the communication matrix $M$ used %in \gdpsgd. (2) We proposed two algorithms \rad and \done to significantly improve the \sg of $M$ and to improve training convergence, under the %same communication traffic cost as the original \gdpsgd. We evaluated our algorithms on both computer vision task and ASR task and demonstrate their %effectiveness. (iii) Our improved system is able to train SWB2000 to reach 7.5\% on SWB, 13.3\% on CH 2.28 hours with 64 V100 GPUs and 7.7\% on SWB, %13.3\% on CH in less than 2 hours with 128 V100 GPUs, on an IBM supercomputer.
\vspace{-0.3cm}

\section{Problem Formulation}
\label{sec:background}

Stochastic gradient descent (SGD) is currently the dominant approach to optimizing deep neural networks. In SGD, models are iteratively updated as shown in Eq.\ref{eqn:sgd}
\begin{align}
    w_{k+1} & = w_{k} - \alpha_{k}\cdot\left[\frac{1}{M}\sum_{m=1}^{M} \nabla f(w_{k};\xi_{k,m})\right]   \label{eqn:sgd}
\end{align}
where $w_{k}$ are the parameters after iteration $k$. The gradient $\nabla f(w_{k};\xi_{k,m})$ is computed using model $w_{k}$ on $M$ randomly drawn data samples indexed by the random variable $\xi_{k,m}$. The $M$ samples form a mini-batch and $M$ is the batch size. $\alpha_{k}$ the learning rate.

In (A)D-PSGD, the weights update rule is given in Eq.\ref{eqn:dsgd}:
\begin{align}
    \mathbf{W}_{k+1} = \mathbf{W}_{k} \cdot \mathbf{T} - \alpha_{k} \cdot g(\mathbf{\Phi}_{k},\bm{\xi}_{k})  \label{eqn:dsgd}
\end{align}
where $\mathbf{W}_{k} = [w^{(1)}_{k}, \dots, w^{(l)}_{k}, \dots, w^{(L)}_{k}]$ is a matrix with each column consisting of model parameters in each learner $l$ at iteration $k$; $\mathbf{T}$ is a doubly stochastic mixing matrix for model averaging among learners given a network topology; $\mathbf{\Phi}_{k} = [\hat{w}^{(1)}_{k}, \dots, \hat{w}^{(l)}_{k}, \dots, \hat{w}^{(L)}_{k}]$ is a matrix with each column consisting of model parameters used for computing gradient in each learner $l$ at iteration $k$. In the asynchronous mode, $\hat{w}^{(l)}_{k}$ may not be equal to $w^{(l)}_{k}$; $\bm{\xi}_{k} = [\xi^{(1)}_{k}, \dots, \xi^{(l)}_{k}, \dots, \xi^{(L)}_{k}]$ is a matrix with each column consisting of indexing random variables for mini-batch samples used for computing gradients in each learner $l$ at iteration $k$ and $g(\mathbf{\Phi}_{k},\bm{\xi}_{k}) = [\frac{1}{M}\sum_{m=1}^{M} \nabla f(\hat{w}_{k};\xi_{k,m}), \dots, \frac{1}{M}\sum_{m=1}^{M} \nabla f(\hat{w}_{k};\xi_{k,m})]$ is a matrix with each column consisting of gradients computed in each learner $l$ at iteration $k$.  In \ssgd, all models are collected and averaged by the total number of $L$ after all learners finish their gradient computation and local model update. The average is then broadcast to each learner. In this case, it can be easily seen that the mixing matrix $\mathbf{T}_{u} = \frac{1}{L}(\mathbf{1}_{L}\trans{\mathbf{1}}_{L})$ where $\mathbf{1}_{L} = \trans{[1,1,\cdots,1]}$. In other words, \ssgd is a special case of \gdpsgd.

%% One of the popular way to carry out distributed learning is through allreduce where local models are collected and averaged by the total number of $L$ after all learners finish their gradient computation and local model update. The average is then broadcast to each learner. In this case, it can be easily seen that the mixing matrix $\mathbf{T}_{u} = \frac{1}{L}(\mathbf{1}_{L}\trans{\mathbf{1}}_{L})$ where $\mathbf{1}_{L} = \trans{[1,1,\cdots,1]}$.

\section{Randomized Mixing}
\label{sec:theory}

The mixing strategy in \ssgd with $\mathbf{T}_{u}$ is fast to reach consensus but it may be communication heavy as models have to be transferred among all learners. One way to reduce the communication cost is to use local averaging. For instance, each learner only averages models  with its left and right neighbors in a ring \cite{icassp19}\cite{interspeech19}. In this case, the mixing matrix is given by
\begin{align}
\mathbf{T}_{0}=\begin{bmatrix}
1/3 & 1/3 &  & & & 1/3\\
1/3& 1/3& 1/3 & & & \\
& 1/3& 1/3 & \ddots& & \\
&	 & \ddots& \ddots &1/3  & \\
&&  &1/3 &1/3 &1/3 \\
1/3&&  & &1/3 &1/3 \\
\end{bmatrix} \label{eqn:fixngbr}
\end{align}
Since each learner only needs to communicate with its immediate neighbors, the communication can be significantly reduced compared to averaging across all learners. It can be shown that, as a doubly stochastic matrix, $\mathbf{T}_{0}$ will converge to $\mathbf{T}_{u}$:
\begin{align}
    \mathbf{T}^{k}_{0} \rightarrow  \mathbf{T}_{u},  \ \ \ \  k \rightarrow \infty .
\end{align}
The speed of convergence is controlled by the spectral gap between the largest (which is always 1) and the second largest eigenvalues of $\mathbf{T}_{0}$. Suppose $\rho=\max\left(|\lambda_{2}(\mathbf{T}_{0})|,|\lambda_{L}(\mathbf{T}_{0})|\right)<1$ is the second largest eigenvalue of $\mathbf{T}_{0}$. We have
\begin{align}
\left\|\mathbf{T}^{k}_{0}-\mathbf{T}_{u}\right\|_2\leq \rho^{k} \label{eqn:rho}.
\end{align}
Given the circulant structure, the eigenvalues of $\mathbf{T}_{0}$ are simply the Fourier transform of the first row. The second largest eigenvalue is given by
\begin{align}
    \rho = \frac{1}{3} + \frac{2}{3}\cos\left(\frac{2\pi}{L}\right)  \label{eqn:trirho}
\end{align}
When $L$ is large, $\rho$ is very close to 1, which indicates a small spectral gap and therefore a slow convergence to consensus.

In this work, we investigate a randomized mixing strategy to accelerate the convergence without increasing the communication cost.  Under this strategy, $L$ learners form a ring and the indices of the learners are randomly shuffled:
\begin{align}
    [1,2,\ldots,L] \rightarrow [\sigma(1), \sigma(2), \ldots, \sigma{(L)}]
\end{align}
where $\sigma(\cdot)$ is a random permutation of the set $\{1,\dots, L\}$. A learner averages models with its left and right neighbors in the mapped indices. The resulting mixing matrix $\mathbf{T}_{\tau}$ of iteration $\tau$ constructed this way is obviously a doubly stochastic matrix and we have
\begin{align}
    \mathbf{T}_{\tau} = \trans{\mathbf{P}}\mathbf{T}_{0}\mathbf{P}   \label{eqn:randperm}
\end{align}
where $\mathbf{P}$ is random permutation matrix. Moreover, we have
\begin{align}
 \widetilde{\mathbf{T}} = \mathbf{E}_{\sigma}[\trans{\mathbf{T}}_{\tau} \mathbf{T}_{\tau}] = \mathbf{E}_{\sigma}[\trans{\mathbf{P}}\trans{\mathbf{T}_{0}}\mathbf{T}_{0}\mathbf{P}]
\end{align}
It can be shown that
\begin{align}
    \widetilde{\mathbf{T}}_{ii} = \frac{1}{3}, \ \ \widetilde{\mathbf{T}}_{ij} = \frac{2}{3(L-1)}, \ i \neq j
\end{align}
It follows that
\begin{align}
	     & \mathbf{E}\left\|\mathbf{T}_{1}\mathbf{T}_{2}\ldots\mathbf{T}_{k}-\mathbf{T}_{u}\right\|_2^2  \nonumber\\
   \leq  & \mathbf{E}\left\|\mathbf{T}_{1}\mathbf{T}_{2}\ldots\mathbf{T}_{k}-\mathbf{T}_{u}\right\|_F^2 \nonumber\\
   =     & \mathbf{E}\left[\text{tr}\left(\left(\mathbf{T}_1\mathbf{T}_{2}\ldots\mathbf{T}_k-\mathbf{T}_{u}\right)^\top\left(\mathbf{T}_1\mathbf{T}_{2}\ldots\mathbf{T}_k-\mathbf{T}_{u}\right)\right)\right] \nonumber \\
   =     & -1 + \text{tr}\left( \tilde{\mathbf{T}}^{k} \right) \nonumber \\
   =     &  (L-1)\left(\frac{1}{3}-\frac{2}{3{(L-1)}}\right)^{k} \leq \frac{L-1}{3^{k}}
\end{align}
which gives
\begin{align}
   \mathbf{E}\left\|\mathbf{T}_1\mathbf{T}_{2}\ldots\mathbf{T}_k-\mathbf{T}_{u}\right\|_2 \leq \frac{\sqrt{L-1}}{(\sqrt{3})^{k}} \label{eqn:randspeed}
\end{align}
Comparing Eq.\ref{eqn:randspeed} with Eqs. \ref{eqn:rho} and \ref{eqn:trirho}, we can see that this randomized mixing strategy converges much faster to consensus than the fixed mixing strategy in Eq.\ref{eqn:fixngbr}.

\section{Design and Implementation}
\label{sec:design}
In each iteration of \textit{\gdpsgd} , a learner calculates gradients in one thread while concurrently exchanging its weights with its left and right neighbors in another thread.

In \textit{\rad},  a learner picks two random neighbors to communicate in each iteration. To achieve this, in each iteration each learner generates a random permutation of all the learner IDs to construct a communication ring (i.e. generates a new mixing matrix $\mathbf{T}$). For this iteration, a learner communicates with the two neighbors in the newly constructed communication ring. We let each learner start with the same random seed to guarantee all learners generate the same random permutation. As in \gdpsgd, each learner sends two messages and receives two messages in each iteration. Assuming all learners are connected with the same communication switch, \rad has the same communication cost as \gdpsgd.

In \textit{\done}, we design the strategy in such way that $\mathbf{W}_{k}\mathbf{T}$ and $g(\mathbf{\Phi}_{k},\bm{\xi}_{k})$ on the RHS of Eq.\ref{eqn:dsgd} are carried out \textit{concurrently}. In addition, the model averaging indicated by $\mathbf{W}_{k}\mathbf{T}$ is realized with allreduce\footnote{An allreduce operation is a reduction operation, which is both associative and commutative such as summation, followed by a broadcast operation. A global summation is an example of an allreduce operation.} divided by $L$.
\begin{align}
    \mathbf{W}_{k}\mathbf{T} = \mathbf{W}_{k}\mathbf{T}_{u}.
\end{align}
On the other hand, the model used for computing the gradients in each learner is the model from the previous round of allreduce
\begin{align}
    \mathbf{\Phi}_{k} = [w^{(1)}_{k-1}, \dots, w^{(l)}_{k-1}, \dots, w^{(L)}_{k-1}]
\end{align}
hence the name delay-by-one decentralized parallel SGD.
The difference between \done and S-PSGD is that S-PSGD requires consensus on gradients before model update, which results in homogeneous models across learners when computing gradients. \done has models updated locally on each learner using different gradients before pushing the models for allreduce across learners, which introduces slight heterogeneity to local models that can be helpful for convergence as demonstrated in our experimental results in \Cref{sec:results}. In contrast, homogeneous models enforced by \ssgd cannot convergence with a large batch size and aggressive learning rate for our ASR task setting\cite{interspeech19}.
A good allreduce implementation can finish each round of communication after effectively 2 messages are sent across the communication network, independent of the number of learners\cite{fsu-allreduce}. We choose the Nvidia NCCL\cite{nccl} as our allreduce implementation. Even though \done has the most favorable \sg (i.e. 1) while incurring the same communication cost as \gdpsgd and \rad, it requires a global synchronization (i.e., allreduce) thus it suffers from the straggler problem in a distributed setting and the communication speed is bounded by the slowest communication link. \Cref{tab:traffic-analysis} summarizes the design choice for each algorithm.
\begin{table*}
  \centering
\begin{tabular}{|c|c|c|c|}
\hline
         & Consensus Convergence*  & Time to Communicate in Each Iteration &  Straggler Avoidance     \\\hline
\gdpsgd   & Slow                               & 2*$\vert M\vert/\vert B\vert$                    & Y              \\ \hline
\rad & Medium                             & 2*$\vert M\vert/\vert B\vert$                    & Y              \\ \hline
\done     & Fast                               & 2*$\vert M\vert/\vert B\vert$                    & N              \\ \hline
\end{tabular}
\caption{Design choice for \gdpsgd, \rad, and \done. We assume a bidirectional network link, in which one learner can send and receive a message at the same time. $\vert M\vert$ means message size, which is the size of neural network. $\vert B\vert$ is the bandwidth between two learners. *For the quantitative consensus convergence analysis, see \Cref{sec:theory}.}
\label{tab:traffic-analysis}
\vspace{-0.5cm}
\end{table*}
\vspace{-0.3cm}

\section{Methodology}
\label{sec:meth}

\subsection{Hardware and Software}

%IBM POWER9 CPUs and NVIDIA Volta GPUs all connected together with NVIDIA’s high-speed NVLink.
%Interconnect Topology: Mellanox EDR 100G InfiniBand, Non-blocking
%The basic building block of Summit is the IBM Power System AC922 node. Each of the approximately 4,600 compute nodes on Summit contains two IBM POWER9 processors and six NVIDIA Volta V100 accelerators and provides a theoretical double-precision capability of approximately 40 TF. Each POWER9 processor is connected via dual NVLINK bricks, each capable of a 25GB/s transfer rate in each direction. Nodes contain 512 GB of DDR4 memory for use by the POWER9 processors and 96 GB of High Bandwidth Memory (HBM2) for use by the accelerators. Additionally, each node has 1.6TB of non-volatile memory that can be used as a burst buffer.

We experiment on an IBM cluster  with a node
architecture similar to the current fastest supercomputer in the world, Summit
\cite{Top500}. This cluster is based on IBM POWER System AC922 nodes with IBM
POWER9 CPUs and NVIDIA Volta V100 GPUs all connected together with NVIDIA’s
high-speed NVLink dual links totaling 50GB/s bandwidth in each direction. Each node contains 22 cores, 512GB of DDR4 memory, 96GB of High Bandwidth Memory (HBM2) for use by the accelerators and is equipped with 6 GPUs.
Nodes are connected with Mellanox EDR 100G Infiniband interconnect technology, each node has a combined network bandwidth of 25GB/s. Each node is equipped with 500GB NVME storage.
We use PyTorch v1.1.0 and IBM Spectrum MPI along with XL compiler suite v16.1.1. For each learner, we use 4 I/O processes to drive the data loading.

\subsection{Models and Dataset}
\label{sec:meth:model}

The hybrid acoustic model used in the SWB2000 experiments is an LSTM with 6 bi-directional layers. Each layer has 1,024 cells (512 cells in each direction). A linear projection layer with 256 hidden units is inserted between the LSTM layers and the softmax layer with 32,000 output units. These 32,000 units correspond to context-dependent hidden Markov model (HMM) states. The LSTM is unrolled with 21 frames and trained with non-overlapping feature subsequences of that length. The feature input is a fusion of 40-dim FMLLR, 100-dim i-Vector and 40-dim logmel with its delta and double delta. The total input dimensionality is 260. The model size is 165MB. The language model is built using publicly available training data from a broad variety of sources. There are 36M 4-grams built on a vocabulary of 85K words. The test set is the Hub5 2000 evaluation set including two parts: 2.1 hours of switchboard (SWB) test set and 1.6 hours of call-home (CH) test set.

Our second benchmark dataset is collection of natural images used as a part of the 2012 edition of the \imagenet Large Scale Visual Recognition Challenge (ILSVRC 2012). The training set is a subset of the hand-labeled \imagenet database and contains 1.2 million images. The validation dataset has 50,000 images. Each image maps to one of the 1000 non-overlapping object categories. The model we use is ResNet-50 (\cite{resnet}).

\section{Experimental Results}
\label{sec:results}
\subsection{Convergence Results}
\begin{figure}[t]%[!htpb]
  \centering

  \subfloat[{SWB2000 heldout loss comparison}]
  {\includegraphics[width=1.0\columnwidth, height=1.2in]{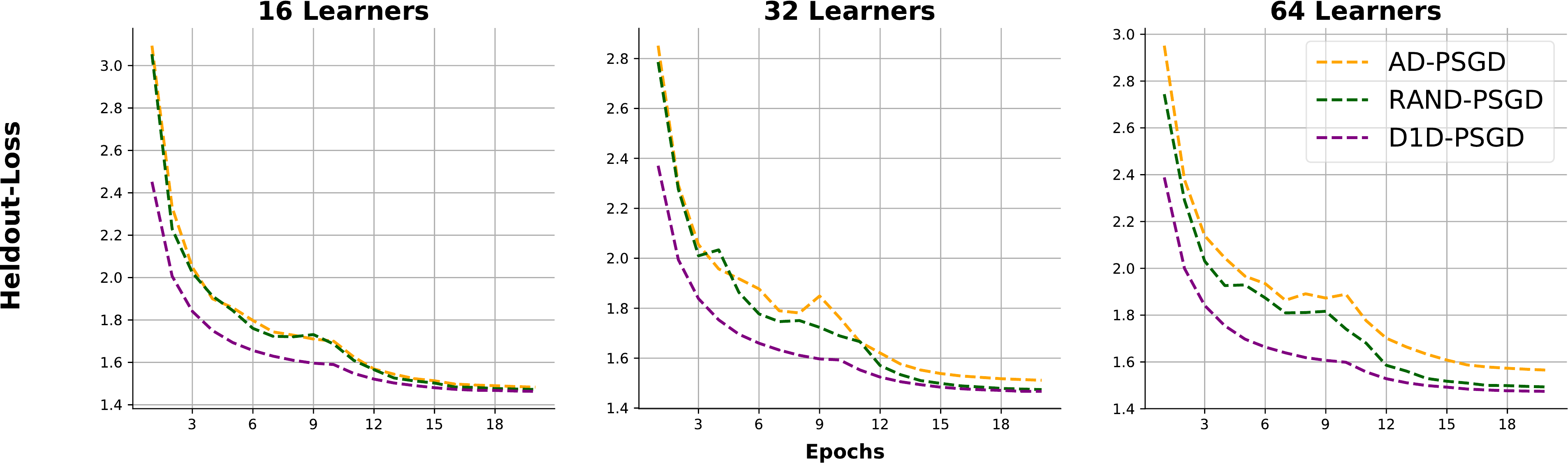}%% \label{fig:slower-link-loss-runtime}
      %\caption{Speedup comparison between \sync (openmpi), \sync(ddl),\hybrid, and \adpsgd. \adpsgd achieves the best speedup (~11X over 16GPUs).}
    \label{fig:heldout-loss-swb2000}
  }
  \\
    \subfloat[{ImageNet top-1 accuracy comparison}]
  {\includegraphics[width=1.0\columnwidth, height=1.2in]{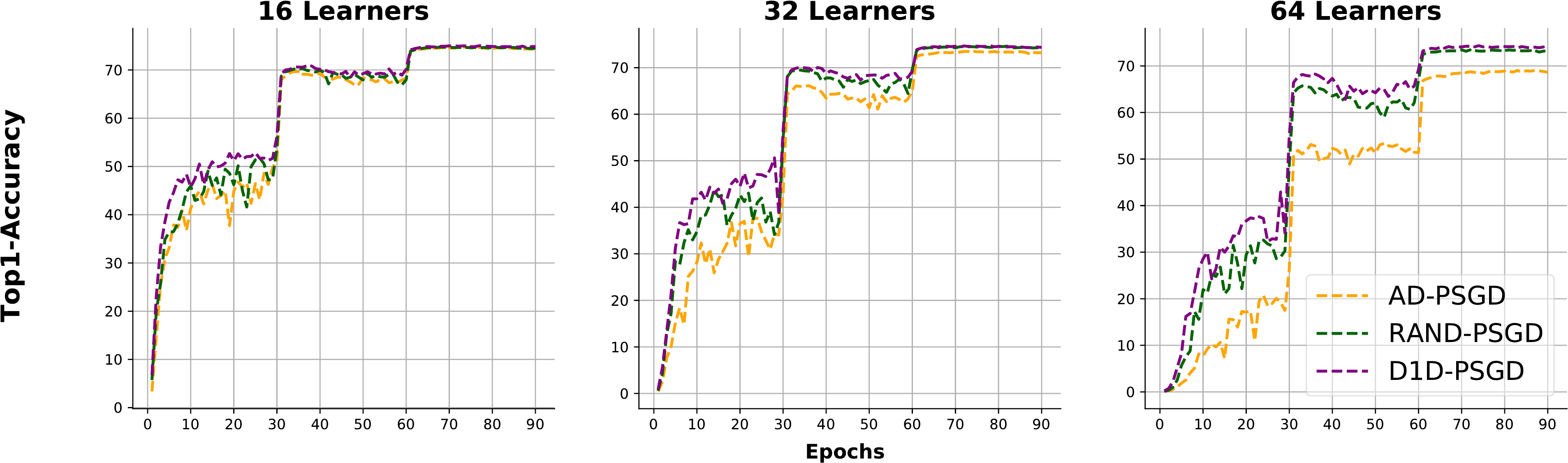}%% \label{fig:slower-link-loss-runtime}
      %\caption{Held-out loss comparison between baseline, \sync, \hybrid, and \adpsgd. They converge similarly.}
    \label{fig:imagenet-accuracy}
  }
  \caption{Convergence comparison of \adpsgd, \rad and \done. \rad and \done converge at a similar speed and both are faster than \adpsgd.}

  \label{fig:conv}
  \vspace{-0.5cm}
\end{figure}
\Cref{fig:conv} illustrates convergence results for \adpsgd, \rad and \done up to 64 learners. For the SWB2000 task, we use the same hyper-parameter settings as described in \cite{interspeech19}, and the total batch size across all the learners is 8192 (e.g., batch size 128 per learner in the 64 learners setting). For the ImageNet task, we use the same hyper-parameter settings as described in \cite{facebook-1hr}, the batch size for each learner is set 32. Up to 64 learners, \rad and \done perform similarly and outperform \adpsgd. \Cref{tab:wer} summarizes the WER of ASR models trained by each algorithm.

\begin{table*}
 \centering
\begin{tabular}{|c|c|c|c|c|c|c|c|c|c|c|}
\hline
\multirow{2}{*}{} & Single Learner& \multicolumn{3}{c|}{16 Learners} & \multicolumn{3}{c|}{32 Learners} & \multicolumn{3}{c|}{64 Learners} \\ \cline{3-11}
                  & Baseline & AD   & RAND   & D1D     & AD   & RAND   & D1D     & AD   & RAND   & D1D     \\ \hline
SWB               & 7.5& 7.6     & 7.6          & 7.4    & 7.9     & 7.7          & 7.6    & 8.1       & 7.8            & 7.5    \\ \hline
CH                & 13.0 & 13.2    & 13.1         & 13.3   & 13.6    & 13.4         & 13.1   & 14.0      & 13.4            & 13.3   \\ \hline
\end{tabular}
\caption{WER comparison after 16 epochs for \adpsgd, \rad, and \done up to 64 learners. AD is short for \adpsgd, RAND is short for \rad, D1D is short for \done. Single learner baseline is trained with batch size 256 under a well-tuned training recipe. For all the other settings, the total batch size is 8192.}
\label{tab:wer}
\centering
\begin{tabular}{|c|c|c|c|c|c|c|c|c|c|}
\hline
\multicolumn{2}{|c|}{16 Learners} & \multicolumn{2}{c|}{32 Learners} & \multicolumn{2}{c|}{64 Learners} &    \multicolumn{2}{c|}{96 Learners}       & \multicolumn{2}{c|}{128 Learners} \\ \hline
WER(SWB/CH)      & Time(hr)      & WER            & Time (hr)      & WER            & Time (hr)      & WER        & Time(hr) & WER            & Time (hr)       \\ \hline
7.4/13.3         & 5.88             & 7.6/13.1       & 3.60              & 7.5/13.3       & 2.28            & 7.7/13.2   & 2.10      & 7.7/13.3       & 1.98            \\ \hline
\end{tabular}
\caption{WER/Time after 16 epochs for \done.}
\label{tab:wer_time}
\vspace{-0.5cm}
\end{table*}

%% \begin{table*}
%%   \centering
%% \begin{tabular}{|c|c|c|c|c|c|c|c|c|c|}
%% \hline
%% \multicolumn{2}{|c|}{16 Learner} & \multicolumn{2}{c|}{32 Learner} & \multicolumn{2}{c|}{64 Learner} & 96 Learner &          & \multicolumn{2}{c|}{128 Learner} \\ \hline
%% WER(SWB/CH)      & Time(hr)      & WER            & Time (hr)      & WER            & Time (hr)      & WER        & Time(hr) & WER            & Time (hr)       \\ \hline
%% 7.4/13.3         & 5.88             & 7.6/13.1       & 3.60              & 7.5/13.3       & 2.28            & 7.7/13.2   & 2.1      & 7.7/13.3       & 1.98            \\ \hline
%% \end{tabular}
%% \caption{WER/Time @16th epoch for \done.}
%% \label{tab:wer_time}
%% \end{table*}

% \usepackage{multirow}

\subsection{Runtime Results}
\begin{figure}%[t]%[!htpb]
  \centering
    {\includegraphics[width=0.6\columnwidth]{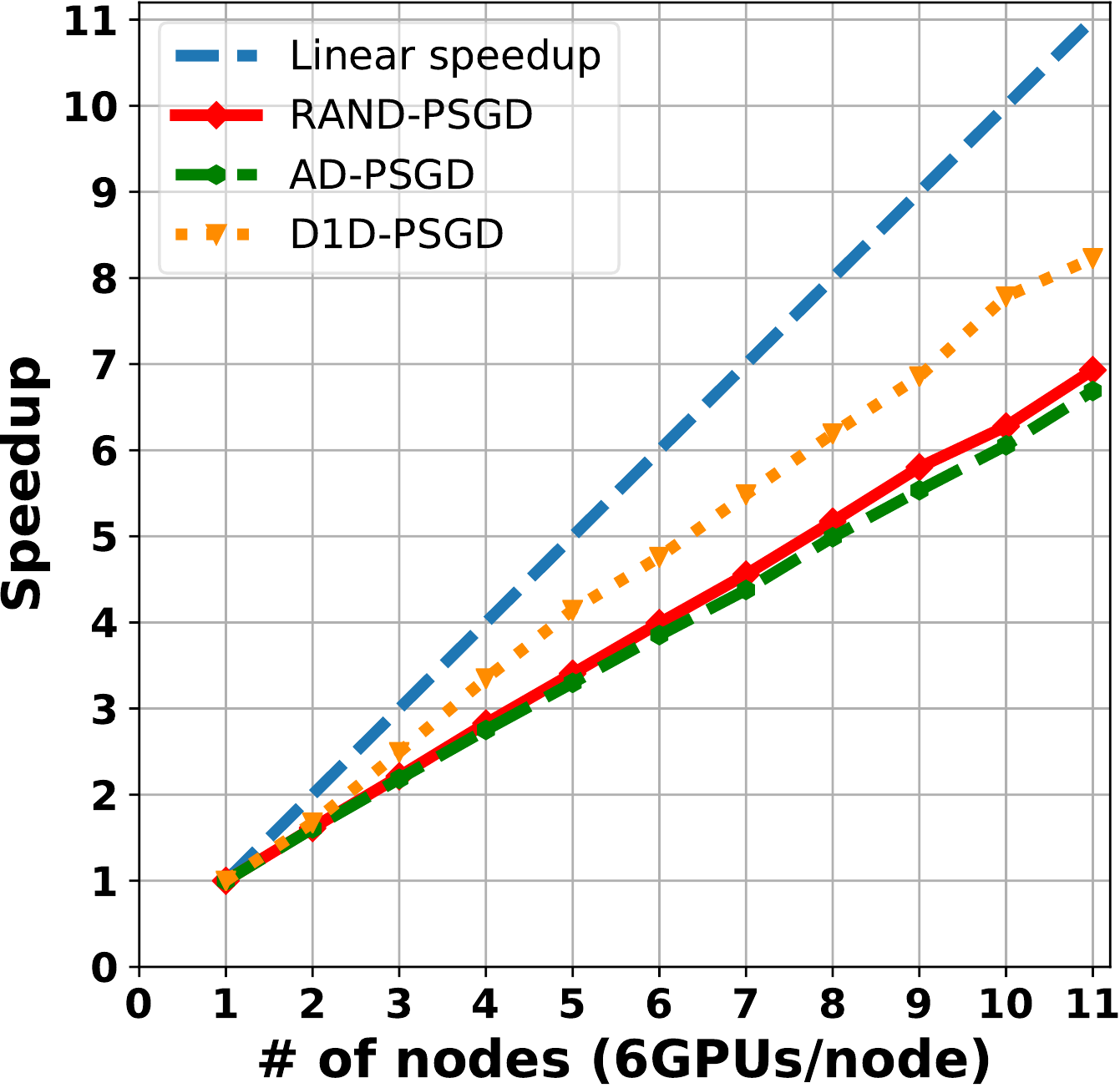}
      \caption{Speedup up to 11 nodes (6 GPUs per node) on IBM POWER9 cluster, batch size per learner is 128. Compared to 1 node runtime, 11 node \done achieves over 8x speedup.}
    \label{fig:SpeedupWSC}
    }
    \vspace{-0.5cm}
\end{figure}

\Cref{fig:SpeedupWSC} shows the speedup up to 11 nodes (i.e. 66 GPUs). \done is the fastest, as NCCL implements sophisticated software pipelining to overlap message exchanges on the bidirectional network link. \adpsgd and \rad run at a similar speed.

%% \done starts with 19.35hr, \adpsgd starts with 32.35hr, \rad stats with 33.53hr. \gdpsgd and \rad achieve similar runtime performance and \done runs the fastest.

\Cref{tab:wer_time} presents the tradeoff of running time and model accuracy using \done. Our system can finish ASR training under 2 hours with 128 GPUs, with slight model accuracy degradation. Note that we keep the total batch size 8192 across all the runs; thus, when there are many learners in the system, the per learner batch size decreases and so does the computation efficiency. Also, when batch size per learner gets smaller, the sample variances per learner gets larger and that might explain the slight model accuracy degradation for 128 learners.

\subsection{Discussion}
In a super-computer environment like ours, where even the slowest communication link bandwidth is 25GB/s and all computing devices are highly homogeneous, \done guarantees the best convergence and can deliver near-metal runtime performance. In a cloud data center environment where network links are usually slow and the straggler problem becomes more prominent, an algorithm built on a global barrier such as allreduce is unlikely to be deployed and we suggest to use \rad as it achieves convergence rate close to \done, has the same traffic cost as \adpsgd and does not rely on global synchronization. Researchers find that \adpsgd outperforms allreduce based algorithms significantly when network links are standard 10Gbit/s ethernet because allreduce speed is bounded by the slowest link \cite{adpsgd, adpsgd-rabbat}.

\section{Related Work}
\label{sec:related}
%% Asynchornous SGD\cite{distbelief,}, once popular large-scale deep learning al%% gorithms has lost is favor, due to inferior model accuracy.
%% Distributed Deep Learning (DDL) is the de facto method to accelerate large-scale deep learning training task. 
Parameter Server\cite{distbelief, adam} based asynchronous parallel SGD approach was first proposed to solve the straggler problems in distributed training. Due to the staleness issue\cite{zhang-ijcai-2016} in the parameter server design, Synchronous Parallel SGD regains its popularity\cite{facebook-1hr,ddl,revisit-sync-sgd}.  
(Asynchronous) Decentralized Parallel SGD is a state of the art distributed deep learning algorithm\cite{dpsgd, adpsgd,adpsgd-rabbat} that removes the need of a centralized parameter server and relax the need for lock-step synchronization. Researchers have recently applied \gdpsgd to training ASR models in record time\cite{icassp19, interspeech19}. One drawback of \gdpsgd is when the number of learners grows, convergence is hampered. To circumvent this problem, \cite{adpsgd}  increases rounds of communication when network links are free, \cite{icassp19}  adapts similar techniques to \done, \cite{interspeech19} groups learners on the same server as one super learner and only applies \gdpsgd among super learners. None of the prior work studies the root cause of this efficiency issue. This paper is the first paper that formalizes the relationship between convergence in \gdpsgd and the number of learners and propose remedies.
Orthogonal to \gdpsgd, many existing works\cite{deepspeech2, bmuf, msr-1bit,seide2014on, amazon-million-hr} adapt a synchronous approach where all learners periodically synchronize and obtain the same set of weights. Among them, \cite{bmuf, amazon-million-hr} reduce the communication frequency by improving the underlying optimizers. %In contrast, \gdpsgd allows learners to have different set of weights at any time.

%% Distributed DL have been applied to speech recognition\cite{deepspeech2,bmuf}, computer vision\cite{facebook-1hr}, language modeling\cite{nvidia-lm-scaling}, and machine translation\cite{fb-mt-scaling} tasks. To reduce the cost of communication, researchers have proposed gradient quantization\cite{msr-1bit,naigang-nips18} and gradient compression\cite{adacomp, terngrad}. All these works adopt a synchronous training method which would become unacceptably slow in a resource-sharing or Cloud environment\cite{gadei}. Asynchronous SGD, based on the parameter-server architecture, is known to have inferior performance and should be avoided when possible\cite{revisit-sync-sgd, facebook-1hr, deepspeech2, zhang2016icdm}. This work is the first that applies Asynchronous Decentralized Parallel SGD (ADPSGD), which has the theoretical guarantee to converge at the same rate as SGD\cite{adpsgd}, to the challenging SWB2000-LSTM task to achieve state-of-the-art model accuracy in a record time. %% without suffering from the straggler problem.
%% Deep learning models are applied to.
%% Distributed deep learning work\cite{dpsgd,adpsgd,gadei,zhang2016icdm,zhang-ijcai-2016}. In this paper we use \cite{adpsgd}. MSR did 1-bit\cite{msr-1bit}. gradient compression\cite{adacomp, terngrad}. Large-scale lstm MT task\cite{fb-mt-scaling}, which says Nvidia lm work is easy. LM scaling , from Nvidia, uses 3X more epochs than baseline\cite{nvidia-lm-scaling}. deep speech \cite{deepspeech}.

\section{Summary}
\label{sec:conclusion}
We identify that in \gdpsgd based training with fixed local model averaging the spectral gap of the mixing matrix decreases when the number of learners grows. This gives rise to slow convergence to consensus and hence decreases the training efficiency. Our proposed algorithms,\rad and \done,  improve the \sg at no extra communication cost compared to \gdpsgd. We show the effectiveness of our proposed techniques on both ASR and computer vision tasks. On an IBM supercomputer, our system is able to train \swbb to reach a WER 7.5\% on SWB and 13.3\% on CH using 64 V100 GPUs in 2.28 hours, and to reach a WER 7.7\% on SWB and 13.3\% on CH using 128 V100 GPUs in 1.98 hours.

\bibliographystyle{IEEEbib}
\bibliography{refs}
\end{document}